\ifdefined\XeTeXversion
\else
  \ifdefined\pdfoutput
    \pdfoutput=1
  \fi
\fi

\documentclass[sigconf, nonacm]{acmart}

\newcommand\preprintcodeurl{https://github.com/sana-ablation/sana-framework}
\newcommand\preprintwebsiteurl{https://sana-ablation.github.io/}
\usepackage{enumitem}
\newcommand{\stitle}[1]{\par\smallskip\noindent\textbf{#1.}\ }

\usepackage{array}
\usepackage{float}
\usepackage{placeins}
\usepackage{multirow}
\usepackage{booktabs}
\usepackage{booktabs}
\usepackage{tabularx}
\usepackage{hyperref}
\usepackage{cleveref}
\usepackage{amsthm}
\newtheorem{example}{Example}
\definecolor{grey}{gray}{0.55}
\definecolor{deltaplus}{HTML}{15803D}   
\definecolor{deltaminus}{HTML}{B91C1C}  
\newcommand{\dgood}[1]{{\footnotesize\color{deltaplus}#1}}
\newcommand{\dbad}[1]{{\footnotesize\color{deltaminus}#1}}

\newcommand{\daplablogo}{%
  \raisebox{-2pt}{\includegraphics[height=1em]{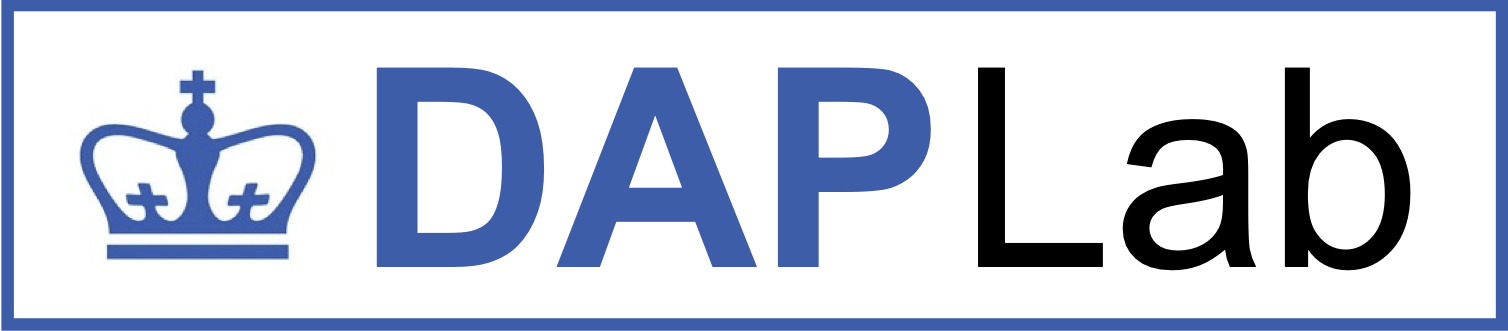}}%
}
\begin{document}
\title{SANA: What Matters for QA Agents over Massive Data Lakes?}

\settopmatter{authorsperrow=1}

\author{Austin Senna Wijaya, Jiaxiang Liu, Haonan Wang, Eugene Wu}
\affiliation{%
  \institution{\daplablogo\ Columbia University}
  \city{New York}
  \country{USA}
}
\email{{asw2215,jl6235,hw2983,ew2493}@columbia.edu}





\begin{abstract}


Exploratory question answering (EQA) over data lakes requires an LLM agent to discover relevant sources, analyze retrieved data, and adapt its actions based on intermediate results. End-to-end accuracy alone cannot distinguish failures in search, planning, data analysis, or the agent's \textbf{Action Policy}: its decisions about what to do next and when to submit an answer. We present SANA 
(Search Agent Navigation Ablation framework), a diagnostic ablation framework that transforms EQA tasks into runtime profiles containing gold source sequence, sanitized subquestions, and execution records. SANA uses these profiles to construct idealized search, planning, and data-analysis tools, allowing each component to be ablated; the residual gap is diagnostic evidence for policy failures.

To illustrate SANA as a reusable evaluation framework, we adapted two recent EQA benchmarks, LakeQA and KramaBench, and evaluated lightweight and mid-sized agents under fixed prompts, budgets, data lakes, and runtimes. Across both benchmarks, data analysis is a consistent bottleneck while planning is less so. Search is a major limitation in LakeQA's large data-lake setting, but less so for the smaller-scale KramaBench. SANA thus deconstructs end-to-end task accuracies into a diagnosis of where data-lake agents fail, and allows for systematic comparisons of progress in search, planning, data analysis, and agent design.

\ifdefempty{\preprintcodeurl}{}{
\par\medskip\noindent\textbf{Artifact Availability:}
Website: \href{\preprintwebsiteurl}{sana-ablation.github.io}. Code and artifacts:
\href{\preprintcodeurl}{sana-ablation/sana-framework}.
}
\end{abstract}

\maketitle

\pagestyle{plain}

\section{Introduction}
Exploratory question answering
(EQA) was recently introduced in LakeQA~\cite{wang2026lakeqaexploratoryqabenchmark} as the problem where an agent must answer questions over facts that are distributed across a large data lake.  The agent must iteratively reason about missing evidence that must be searched for, inspect and analyze retrieved data, and use intermediate findings to guide subsequent actions. This models practical analytics over data lakes: data scientists and business analysts can formulate high-level questions, but not which tables, metadata records, filters, or computations are needed to answer it.

Evaluating EQA is difficult because success depends not only on the quality of individual capabilities (e.g., search, planning, analysis of retrieved datasets), but also on how the agent composes them over a long trajectory. We refer to the agent's \textbf{Action Policy}, or simply \textbf{Policy}, as the turn-level decision process that selects the next action given the question, discovered sources, intermediate observations, tool errors, and remaining budget. At each turn, the policy must decide whether to search, inspect, compute, validate, revise, or submit an answer.


Prior work has advanced individual capabilities needed for EQA, including agentic retrieval~\cite{balaka2025pneuma, zhang2026autoddg, li2026semanticsimilarityrethinkingretrieval}, planning and decomposition~\cite{pourreza2023dinsql, min2019multi}, tool use~\cite{qin2024toolllm, schick2023toolformer}, and code/SQL generation~\cite{yu2018spider, lei2025spider, li2023can}. EQA requires these capabilities to interact in a single long-horizon loop~\cite{wang2026lakeqaexploratoryqabenchmark}. Recent evaluations study facets of EQA: DCI~\cite{li2026semanticsimilarityrethinkingretrieval} gives agents direct grep-style corpus access, while Metadata Reasoner~\cite{zhang2026agenticapproachmetadatareasoning} evaluates sufficient and minimal dataset selection. Neither isolates the contributions of search, planning, analysis execution, and agent policy to end-to-end accuracy. Consequently, failures from missing data, ignored evidence, incorrect decomposition, and brittle SQL or code remain conflated. SANA provides controlled ablations that diagnose these bottlenecks in end-to-end EQA agents.

\begin{figure}[t]
  \centering
  
  \includegraphics[width=1.0\linewidth,keepaspectratio]{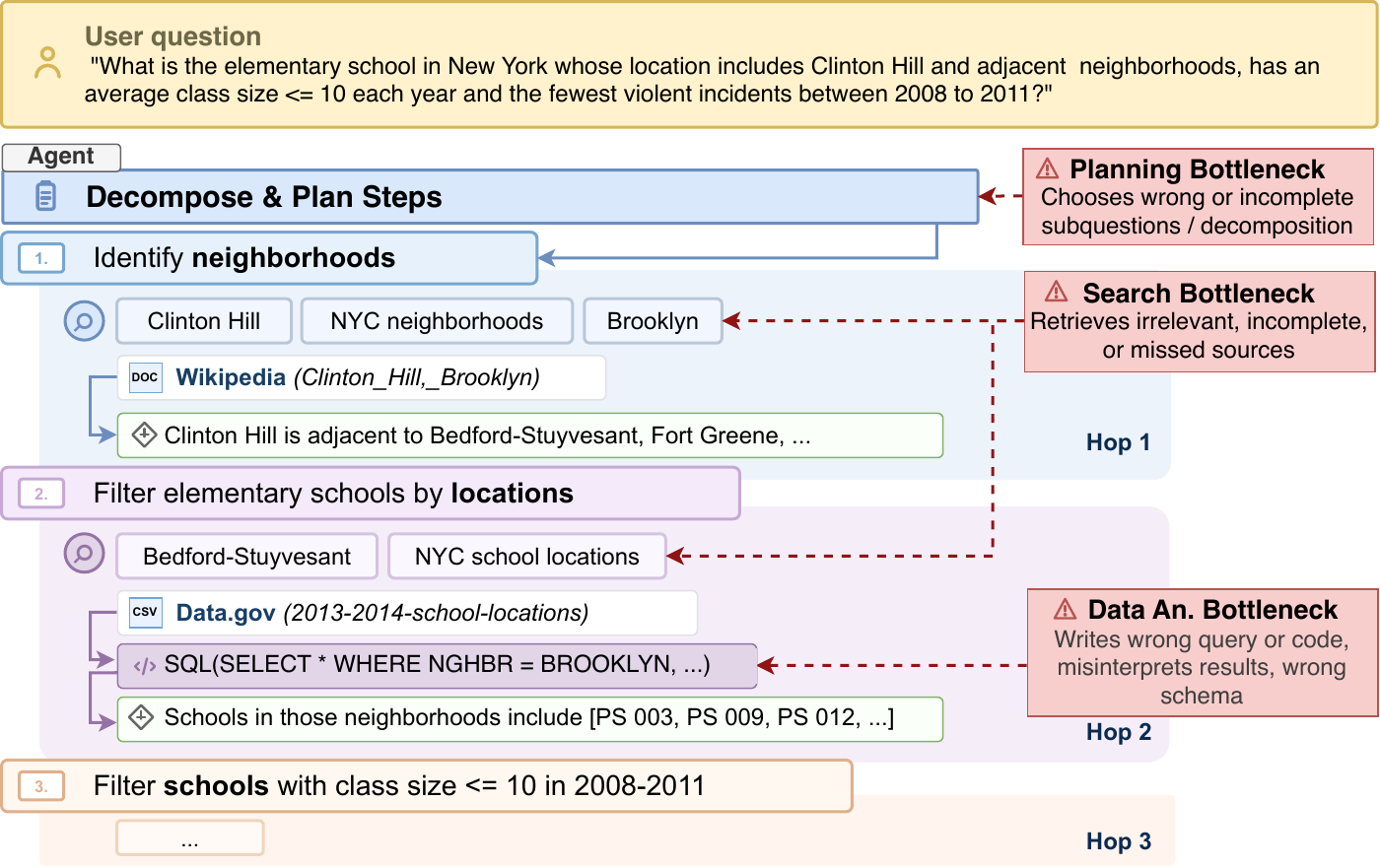}
  \caption{EQA agent runtime loop, with inherent bottlenecks during planning, searching, and data analysis.}
  \Description{A workflow diagram showing user question decomposition, agent memory, policy, search, candidate routing, text and table source processing, observation updates, and final answer generation. 
  }
  \label{fig:sana-workflow}
\end{figure}

To bridge this gap, we propose SANA 
(Search Agent Navigation Ablation framework), an ablation framework to study EQA agent policies. Specifically, we develop an idealized EQA agent whose trajectory planning, search system, and data analysis code are designed using the ground truth. This produces an upper bound for those components while preserving the agent's policy decisions, such as invoking tools and submitting its final answer. We then systematically ablate each component with widely used implementations (e.g., BM25 or hybrid for search).

\begin{figure}[t]
    \centering
    \includegraphics[width=\linewidth,keepaspectratio]{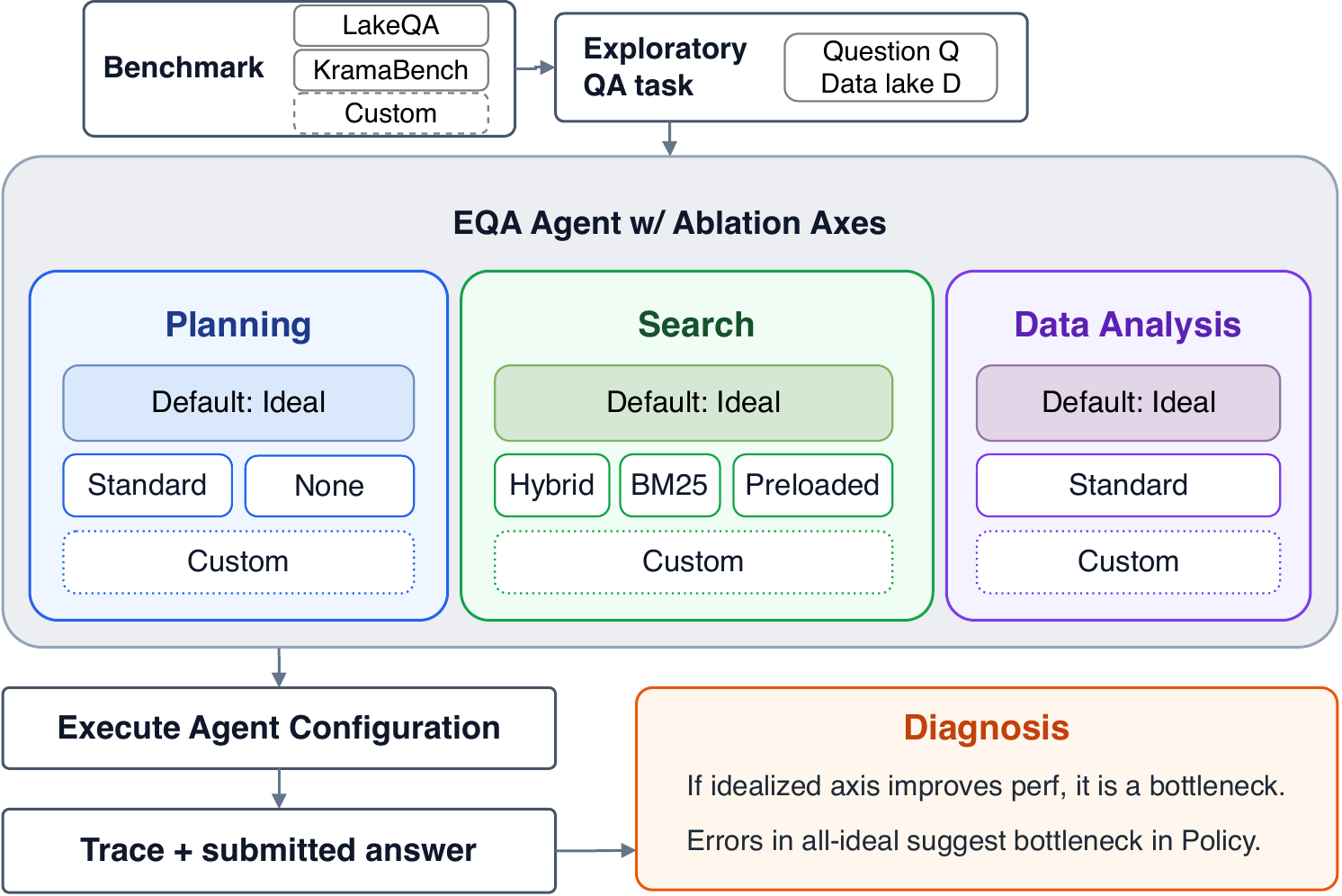}
    \caption{SANA's three runtime components as an ablation study. Each component can be evaluated with idealized, standard, baseline, or custom implementations. }
    \label{fig:sana-ablation-overview}
\end{figure}


By idealizing individual components, SANA distinguishes failures caused by planning, search, or data-analysis execution from failures that remain in the agent policy. These residual policy failures include pursuing the wrong source, failing to validate intermediate evidence, submitting an incorrect final answer despite useful evidence, or exhausting the budget without progress. They expose concrete targets for future EQA agents, including stronger evidence tracking, validation, final-answer checking, and stopping criteria.

To summarize, we make the following contributions:

\begin{enumerate}[label=(\arabic*), leftmargin=*, itemsep=0pt]
\item \textbf{SANA}, a diagnostic ablation framework that converts solved EQA tasks into runtime profiles and uses them to construct idealized planning, search, and data-analysis interfaces while preserving the agent policy.

\item A \textbf{controlled evaluation} on LakeQA~\cite{wang2026lakeqaexploratoryqabenchmark} and a conversion of KramaBench~\cite{lai2025kramabench} that isolates the end-to-end contribution of planning, search, and data-analysis execution under fixed agents, prompts, budgets, data lakes, and runtimes.

\item An \textbf{empirical diagnosis of EQA bottlenecks}: data-analysis execution is a consistent limitation across both benchmarks; search becomes a major limitation in LakeQA's larger-scale discovery setting; and residual agent-policy failures remain even after component idealization.
\end{enumerate}

%

\section{Related Work}
\stitle{Benchmarks for exploratory QA}
QA benchmarks test multi-hop reasoning over heterogeneous data types,
including HotpotQA \cite{yang2018hotpotqa}, MuSiQue~\cite{trivedi2022musique},
OTT-QA~\cite{chen2020open}, TAT-QA~\cite{zhu-etal-2021-tat}, and
FeTaQA~\cite{nan2022fetaqa}. These benchmarks stress reasoning over documents, tables, or mixed evidence, but typically operate over a fixed corpus or compact evidence pool. Data-lake benchmarks move closer to
exploratory analysis: LakeQA~\cite{wang2026lakeqaexploratoryqabenchmark} requires agents to
search, inspect, compute, and synthesize over large data lakes, while
KramaBench~\cite{lai2025kramabench} evaluates data-to-insight pipelines
over multiple structured and unstructured sources. Recent agentic
data-discovery evaluations such as
DCI~\cite{li2026semanticsimilarityrethinkingretrieval} and Metadata
Reasoner~\cite{zhang2026agenticapproachmetadatareasoning} also study
agents that interact with a corpus or select task-relevant datasets, but they do not isolate search, planning and execution as separate failure modes. SANA builds on this benchmark direction, but
uses these tasks to diagnose which part of an EQA agent fails.

\stitle{Component methods for data-lake QA}
Prior work improves individual components needed for EQA. Dataset
discovery systems such as D3L~\cite{bogatu2020dataset} and
Starmie~\cite{fan2022semantics} retrieve joinable, unionable, or
semantically related tables, while Pneuma~\cite{balaka2025pneuma} and
AutoDDG~\cite{zhang2026autoddg} use LLM-generated descriptions to
improve tabular retrieval. Text-to-SQL and data-agent benchmarks such as
BIRD~\cite{li2023can} and Spider 2.0~\cite{lei2025spider} evaluate
data analysis execution once the relevant database context is known. Decomposition
methods such as DecompRC~\cite{min2019multi} and
DIN-SQL~\cite{pourreza2023dinsql} expose intermediate structure for
multi-hop reasoning or SQL generation. SANA treats these as interacting
bottlenecks rather than isolated tasks: decomposition must be grounded and guide the agent's next discovery and analysis, discovery must feed analysis, and
analysis must feed later discovery.

\stitle{Agentic RAG and navigation}
Agentic RAG systems use an LLM-controlled policy to decide when to
retrieve, call tools, update context, and stop~\cite{singh2025agentic},
building on reasoning-and-acting methods such as
ReAct~\cite{yao2022react}, IRCoT~\cite{trivedi2023interleaving}, and
Self-RAG~\cite{asai2024self}. Recent systems such as
MA-RAG~\cite{nguyen2025ma} and A-RAG~\cite{du2026rag} further explore
multi-agent or hierarchical retrieval architectures. These systems expose
the same navigation decisions that arise in EQA, but their evaluations
often conflate model policy, retrieval infrastructure, data profiling,
tool design, and execution errors. SANA provides a controlled evaluation
substrate by holding the task and runtime fixed while ablating search,
planning, and execution.

\section{The SANA Ablation Framework}
\label{sec:sana-framework}
SANA is an ablation framework for search, planning, and execution---the three major components in EQA.     This section introduces EQA tasks, why we focus on these components, and the SANA design. 


\subsection{Anatomy of an Exploratory QA Task}
\label{subsec:task-representation}

Here, we describe the structure of an EQA task and inter-dependencies between data discovery, question decomposition, data analysis, and tool implementation that are needed to successfully answer a task. In short, data discovery must retrieve the necessary datasets, but if it retrieves too many, then it makes question decomposition challenging.   Similarly, if the data analysis tools are poorly implemented or not sufficiently expressive, then the agent cannot extract the necessary information from the datasets to determine answers or next steps.

\subsubsection{An EQA Task.}
In LakeQA~\cite{wang2026lakeqaexploratoryqabenchmark}, an exploratory question-answering (EQA) task is defined as
$
\tau = (Q, \mathcal{D}, \mathcal{D}_{\mathrm{gold}}, \mathcal{T}, A^\star, B),
$
where \(Q\) is a natural-language question, \(\mathcal{D}=\{d_1,\ldots,d_N\}\) is a large data lake containing structured and unstructured datasets, \(\mathcal{T}\) is the set of tools available to an LLM agent, \(A^\star\) is the gold answer to \(Q\), \(\mathcal{D}_{\mathrm{gold}}\) contains a minimal set of datasets\footnote{In practice, $\mathcal{D}_{\mathrm{gold}}$ may not be unique, as it's currently intractable to enumerate all minimal sets.} sufficient to derive the gold answer, and the budget \(B\) is the maximum number of tool calls permitted.

Generating an answer $\hat{A}$ for an EQA task requires decomposing \(Q\) into a sequence of subquestions \((Q_1,Q_2,\ldots,Q_K)\), discovering each subquestion's gold dataset \((d_{\mathrm{gold}}^1,\ldots,d_{\mathrm{gold}}^K)\), and answering \((A_1,A_2,\ldots,A_K)\), such that each intermediate answer can be derived from the current subquestion and gold dataset, and the previously derived answers:
\begin{equation}
Q_i, d_{\mathrm{gold}}^i, A_1,\ldots,A_{i-1} \vDash A_i,
\quad i=1,\ldots,K.
\label{eq:eqa-step-entailment}
\end{equation}

Operationally, an agent must use the tools in $\mathcal{T}$ to discover candidate datasets, analyze them to assess their task-relevance, extract information to compute intermediate answers, and derive the final answer $\hat{A}$---all within the tool call budget.  
We next describe the classes of tools in \(\mathcal{T}\) and the main challenges posed by EQA.  Here, $t$ denotes the $t^{th}$ tool call made by the agent.  

\subsubsection{Agent tools}
The tool set \(\mathcal{T}\) supports two classes of operations: (i) dataset search \(f(q, k)\), over the data lake and (ii) data analysis \(g(c,d)\) over a dataset $d\in \mathcal{D}$. The search tool \(f(q,k)\) currently is based on keyword search over dataset contents; it takes as input keywords \(q\), and returns the top-\(k\) documents from the data lake. Analysis \(g(c,d)\) takes as input an executable program (e.g., Python program, SQL query) \(c\) to evaluate over dataset \(d_i\), and returns the corresponding execution I/O traces. 

We use \(f_t=f_t(q_t,k_t)\) and \(g_t=g_t(c_t,d_t)\) as shorthand to denote the specific tool and its arguments in the \(t^{\mathrm{th}}\) tool call. We denote the cumulative set of discovered datasets at turn \(t\) as \(\mathbf{D}_{\mathrm{disc}}^{t}\).  In practice, an agent can analyze multiple datasets together, but the text refers to a single dataset for legibility.  


\subsubsection{Planning.}
\label{subsubsec:question-decomposition}
The agent must decompose the EQA question by iteratively identifying the next subquestion to answer given the previous evidence.  Formally, given the current discovered datasets \(\mathbf{D}_{\mathrm{disc}}^{t}\) at tool call $t$, planning produces an ordered sequence of subquestions and evidence choices
$$
\bigl((\widehat{Q}_1,\widehat{d}_1),(\widehat{Q}_2,\widehat{d}_2),\ldots\bigr)
$$
where \(\widehat{d}_i\) may be unknown at the time \(\widehat{Q}_i\) is first formulated and must be discovered through subsequent search calls. 

\subsubsection{Search.}
\label{subsubsec:dataset-discovery}
The EQA tasks are designed such that the gold evidence $\mathcal{D}_{\mathrm{gold}}$ is necessary in order to correctly answer the subquestions and the task $Q$.  
Thus, the agent must, at a minimum, within a budget of $B$ tool calls, identify a superset  $\mathcal{D}_{\mathrm{disc}}^{B} \supseteq \mathcal{D}_{\mathrm{gold}}$. However, high recall alone is not sufficient: \(\mathcal{D}_{\mathrm{disc}}^{B}\) could degenerate to the entire data lake. Search precision is therefore also critical, because irrelevant datasets increase the burden on downstream planning and analysis.

\subsubsection{Data analysis.}
\label{subsubsec:insight-extraction}
Even if the agent retrieves the gold dataset for the current subquestion, it still needs to analyze that dataset to derive the intermediate answer. In short, the agent must correctly implement the entailment in \cref{eq:eqa-step-entailment} to derive $\widehat{A}_i=g_t(\widehat{c}_i,\widehat{d}_i)$ without execution failure.    If \(\widehat{A}_i = A_i\) then the agent has correctly answered subquestion $i$.


\subsubsection{Residual Failures.}
The three challenges above define the major failure classes targeted by SANA's ablations. However, they do not exhaust all ways an EQA task can fail. For example, even if the agent correctly derived the answers to the correct query decomposition, it may still hallucinate an incorrect final answer \(\widehat{A} \neq A^\star\).  They may also waste the budget by completing irrelevant actions, skipping necessary steps, or hallucinating.  We discuss these residual failures in Section~\ref{subsec:failure-analysis}.

\subsection{SANA Ablation}
\label{subsec:ablation-axes}

SANA isolates the effect of the major components in EQA (\cref{subsec:task-representation}) by ablating search, planning, and data analysis in isolation or combination.  Ablation replaces a component with an idealized version or other implementations.  We first describe why ablation cannot naively swap out components and needs to be {\it intent-based}, then describe the task profile that we use to generate idealized versions for each ablated component.  We finally describe the ablation methods for each component and our implementation for each.

\subsubsection{Intent-based Ablation}
Isolating the contribution of the system components alone requires careful design.  For instance, a naive approach to ablate search is to simply give $\mathcal{D}_{\mathrm{gold}}$ to the agent.  However, the agent should still be expected to formulate the appropriate search query; we aim to control only the search tool's quality.  

SANA therefore extracts the semantic intention \(a_t\) as natural language behind each tool call. For search, SANA extracts \(a_t\) as the keyword $q$; for execution, SANA extracts \(a_t\) as the analysis goal that $c$ seeks to achieve on $D$.

\subsubsection{Task Profiles}

\begin{figure}[t]
  \centering
  \includegraphics[width=0.92\linewidth]{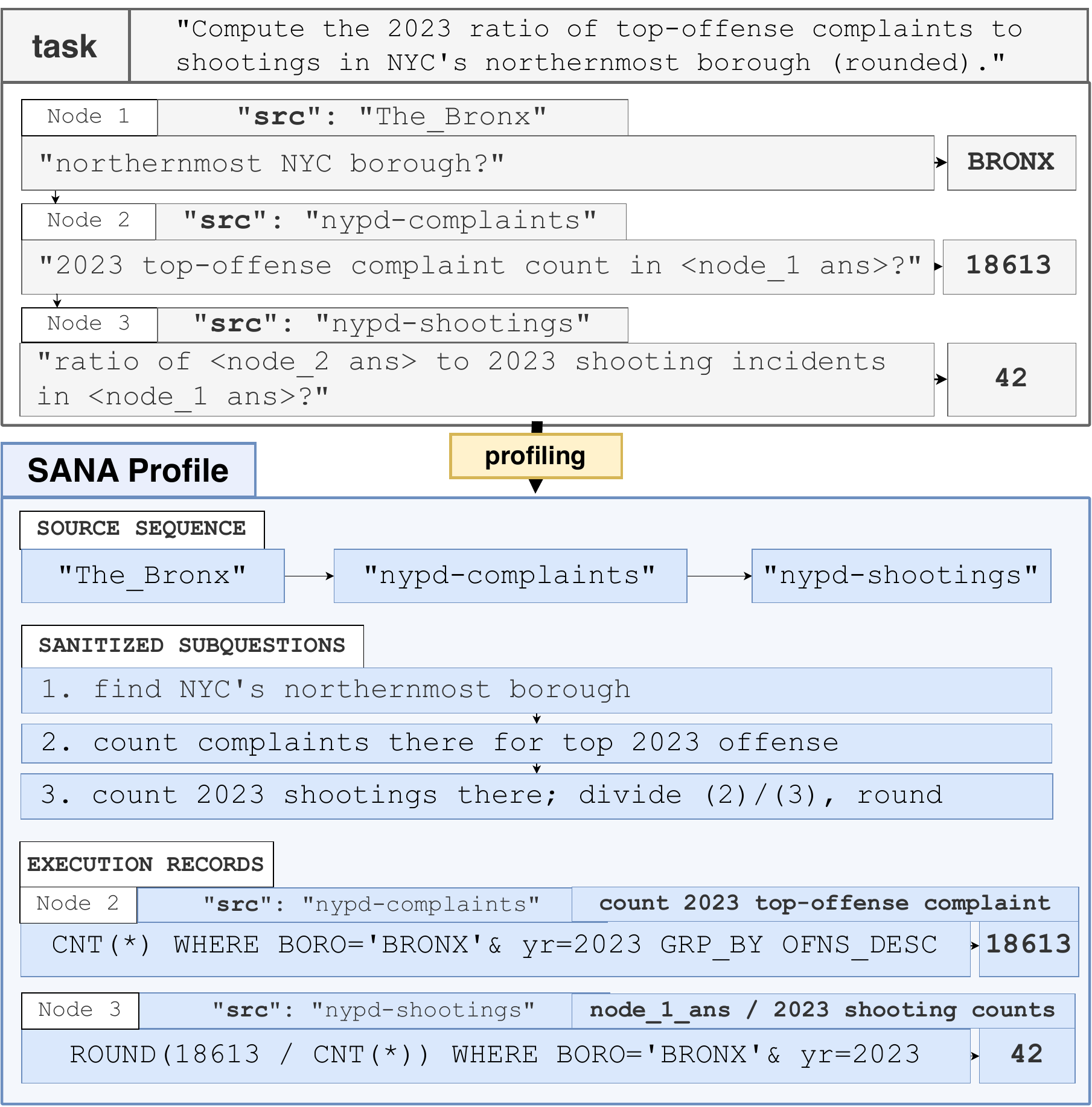}
  \caption{SANA annotates a task into source sequence, sanitized subquestions, and execution records. }
  \label{fig:task-plan-mirror}
\end{figure}

SANA constructs task profiles that contain ground-truth information, and uses them to synthesize 
idealized  implementations of each ablated component (\Cref{fig:task-plan-mirror}). 
A profile is defined by: (Source Sequence) the sequence of gold datasets $\mathcal{D}_{\mathrm{gold}}$ needed for each subquestion,
(Sanitized Subquestions) the sequence of subquestions $\widetilde{\mathcal{Q}}$ that have been sanitized to not leak information about the gold dataset, and (Execution Records) that specify the dataset $d$, the query $c$, the correct intent for analysis over the dataset $a$, and the answer $A$ for the subquestions that require data analysis.  

\begin{example}
  \Cref{fig:task-plan-mirror} adapts a LakeQA task.  In addition to explicitly listing the order of the three gold datasets, it removes direct mentions of dataset names in each subquestion to avoid leaking search targets. For example, a mention of \texttt{nypd\_crimes} is rewritten as ``find crime reports reported by NYPD.''   Subquestion 1 simply retrieves \texttt{The Bronx} dataset so it does not have an execution record, while the latter steps do have analysis records. The first execution record states the dataset is \texttt{nypd-complaints}, the correct analysis intent is to count the top-offense complaints in 2023, the correct SQL, and the answer \texttt{18613}.  
    
\end{example}

In SANA, we adapted tasks from LakeQA and KramaBench.  The former contains all of the necessary information for the profile, and we use an LLM to remove identifying signals from each subquestion. 
The latter only has a question and a monolithic solution script that executes on the task's dataset, so we use an LLM to decompose this solution script into LakeQA-like sequential, dependent hops such that each hop's code is execution-verified against the intermediate answer and no subquestion spans more than one source.

\subsubsection{Planning Ablation.}
\label{subsubsec:planning-ablation}
The planning ablation isolates errors in question decomposition by providing the correct sequence of evidence needs while preserving the agent's responsibility to formulate search queries, conduct data analysis, and synthesize the final answer. In the standard setting, stronger planners produce sequenced subquestion--dataset pairs closer to the annotated sequence \(\bigl((Q_1,d_{\mathrm{gold}}^1),\ldots,(Q_K,d_{\mathrm{gold}}^K)\bigr)\).


To generate the idealized component, we provide the agent with $\widetilde{\mathcal{Q}}$ and ask it to more explicitly state the goal, suggest the tool type to use (e.g., search, query), fallback hints in case a tool call fails. For instance, ``count complaints for that borough's top 2023 offense'' would be rewritten as ``{\it extract} top-{\it count} records of the borough in 2023 in the {\it NYC complaints dataset}.''



\subsubsection{Search Ablation.}
\label{subsubsec:search-ablation}
The search ablation isolates the effect of search precision: it removes irrelevant retrieval results while preserving the agent's responsibility for deciding what to search for, when to search, and when sufficient evidence has been collected. In the standard setting, the agent searches over the full data lake \(\mathcal{D}\), so even when the search tool retrieves gold datasets, it may also return irrelevant datasets that increase downstream planning and analysis burden. To construct the ideal search setting, SANA replaces the standard search tool with an oracle-like operator \(f^{\mathrm{ideal}}\) whose search space is restricted to \(\mathcal{D}_{\mathrm{gold}}\). Thus, every returned dataset is relevant to the task:
\[
f^{\mathrm{ideal}}(q,k) \subseteq \mathcal{D}_{\mathrm{gold}}
\qquad \forall q,k .
\]

We implement idealized search by providing the set of gold datasets, the search query, and the search intent to a subagent that selects the datasets that match the search intent. Each dataset is augmented with metadata: an LLM-generated description~\cite{balaka2025pneuma} and a content preview.  For a table, we use its schema and row samples; for a document, we use a 100-word preview.   
In \Cref{fig:sana_runtime_operators}, ``NYC complaints 2023'' would return \texttt{nypd-complaints}, while ``NYC farmers market'' returns \texttt{dataset not found}.

%
%

\subsubsection{Data Analysis Ablation.}
\label{subsubsec:data-analysis-ablation}
The data analysis ablation isolates implementation errors in executing the intended computation. In the standard setting, given a subquestion--dataset pair \((Q_i,d_i)\), stronger agents are more likely to construct an analysis query that correctly derives the annotated intermediate answer \(A_i\).


SANA replaces the standard analysis tool with \(g^{\mathrm{ideal}}_t(c_t, d_t, a_t)\), which contains an additional parameter: the agent's intent. If $a_t$ matches with the annotated entailment \(a_i\) for each $(Q_i, d_{\mathrm{gold}}^i)$ pair, SANA returns the verified intermediate answer:
$
g^{\mathrm{ideal}}(c_t,d_t) = A_i
$, otherwise, SANA infers the computational intent from $a_t$ and ensures $c_t$ correctly implements it without execution failure.

\begin{figure}[h]
    \centering
    \includegraphics[width=\linewidth,keepaspectratio]{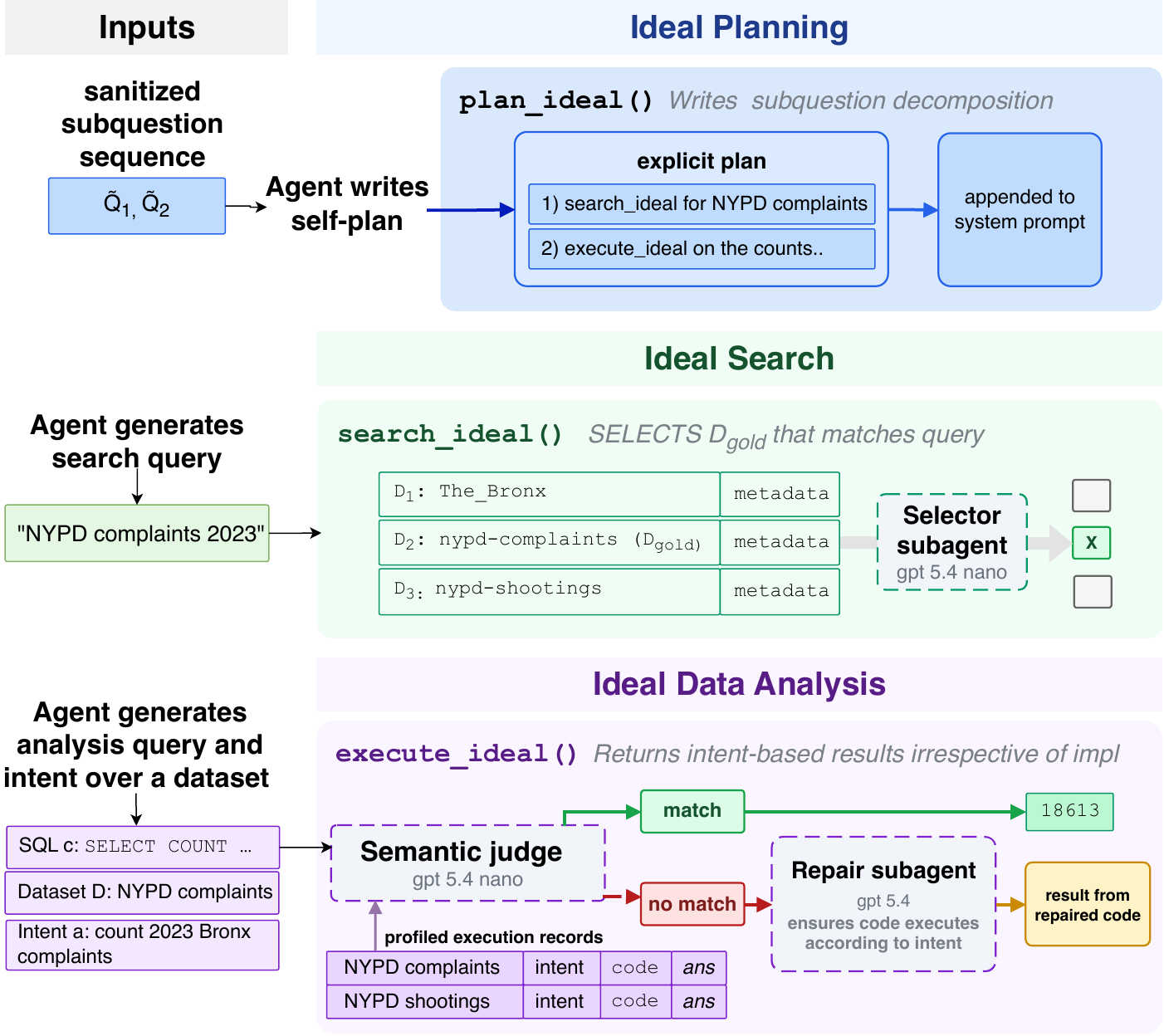}
    \caption{SANA's ideal ablation consumes one profile input and is modularized as a runtime tool.
    }
    \label{fig:sana_runtime_operators}
\end{figure}

We implement idealized data analysis (\Cref{fig:sana_runtime_operators}) by first checking if $d_t$ is the gold dataset. If the agent's intent is the same as the dataset's analysis intent in the execution record (via a subagent), we return the gold answer. If the intents don't match or \(d_t\) is not a gold dataset, SANA uses a stronger model (\texttt{gpt-5.4}) to generate code that matches the intent and returns the result. If the code execution fails, the model repairs the code and tries again for at most two iterations.   

%
%


\section{Evaluation Setup}
This section describes the task suites, ablation conditions, execution conditions, and metrics used to evaluate SANA. We follow the LakeQA evaluation setting with a budget of \(B=30\) tool calls per run and a maximum runtime of 600 seconds; each run starts from a natural-language question and ends when the agent either calls the answer-submission tool or reaches the run budget. We evaluate two model settings: \texttt{gpt-5.4-nano} as the weaker agent and \texttt{gpt-5-mini} as the stronger agent. 

\subsection{Benchmarks and Tasks}
\label{subsec:benchmarks}

We evaluate on two task suites. The first is the LakeQA
\texttt{tasks\_mini} subset~\cite{wang2026lakeqaexploratoryqabenchmark}, containing 135 EQA tasks. The second is the converted KramaBench~\cite{lai2025kramabench}. 
To make KramaBench require data discovery, we expose its dataset as a collection of sources~\cite{zhang2026agenticapproachmetadatareasoning}; to align with the LakeQA evaluations with \(B=30\) tool calls, we remove the Kramabench tasks with \(|\mathcal{D}_{\mathrm{gold}}|\geq 20\) since these tasks require excessive data discovery relative to the run budget--this reduces the task count from 104 to 83.

\begin{table}[h]
\caption{Benchmark statistics for SANA evaluations. \textbf{Sources} is the number of distinct gold sources across all tasks }
\centering
\small
\setlength{\tabcolsep}{3pt}
\label{tab:benchmark-stats}
\begin{tabular}{lrrrr}
\textbf{Benchmark} & \textbf{Tasks} & \textbf{Sources} & \textbf{Avg. $|D_{\mathrm{gold}}|$} & \textbf{Lake Size $|\mathcal{D}|$}\\
\midrule
LakeQA & 135 & 499 & 6.9 & $\sim$40 million\\
KramaBench-conv. & 83 & 187 & 2.3 & 1764 \\
\end{tabular}
\end{table}

\subsection{Ablation Conditions}
\label{subsec:ablation-runs}

\stitle{Targeted Ablations}Each SANA condition is a tuple of planning, search, and data-analysis
modes, as summarized in Table~\ref{tab:sana-conditions}. The targeted ablations vary one axis at a time while holding the other axes idealized. For example, when ablating planning, search and data analysis are kept idealized such that there are no conflicting bottlenecks. 

\begin{table}[h]
\centering
\footnotesize
\setlength{\tabcolsep}{4pt}
\renewcommand{\arraystretch}{1.08}
\caption{SANA condition space. Each run selects one mode from each axis.}
\label{tab:sana-conditions}
\begin{tabularx}{\columnwidth}{@{}llX@{}}
Axis & Mode & Description \\
\midrule

\multirow[t]{3}{*}{Planning}
& \textsc{Naive}
& No planning tool; the agent answers directly from the question and available tools. \\

& \textsc{Standard}
& Uses the planning tool, but with no sanitized subquestion sequence; the plan is entirely self-written. \\

& \textsc{Ideal}
& Receives the sanitized subquestion sequence and uses the planning tool to store a plan derived from the reasoning chain. \\

\midrule

\multirow[t]{4}{*}{Search}
& \textsc{Naive}
& BM25 sparse lexical search over the data lake. \\

& \textsc{Standard}
& Hybrid search with Reciprocal Rank Fusion (RRF), with LLM-generated table descriptions following Pneuma~\cite{balaka2025pneuma} and AutoDDG~\cite{zhang2026autoddg}. \\

& \textsc{Ideal}
& Ideal search tool $f^{\mathrm{ideal}}(q,k)$ over \(\mathcal{D}_{gold}\). \\

& \textsc{Preloaded}
& \(\mathcal{D}_{gold}\) placed directly in agent context; no search tool. \\

\midrule

\multirow[t]{2}{*}{Data analysis}
& \textsc{Standard}
& Writes and runs SQL or Python through the ordinary execution tools. \\

& \textsc{Ideal}
& Ideal execution tool \(g^{\mathrm{ideal}}_t(c_t, d_t, a_t)\) for data analysis. \\

\end{tabularx}
\end{table}


\stitle{End-to-end ablations} From \cref{tab:sana-conditions}, we also define three end-to-end conditions:
(i) \textsc{Naive} denotes \textsc{Naive} Planning, \textsc{Naive} Search, and \textsc{Standard} data analysis; 
(ii) \textsc{Standard} denotes \textsc{Standard} planning, \textsc{Standard} search, and \textsc{Standard} data analysis; 
(iii) \textsc{Ideal} denotes \textsc{Ideal} planning, \textsc{Ideal} search, and \textsc{Ideal} data analysis. 
This comparison measures the improvement from the lower-bound baseline (\textsc{Naive}) to a stronger non-ideal implementation (\textsc{Standard}), as well as the remaining gap to the idealized upper bound (\textsc{Ideal}).




\subsection{Execution Conditions}
\label{subsec:execution-condition}

We vary the conditions in \cref{subsec:ablation-runs} while holding the execution environment below fixed.


 
\stitle{Execution environment} 
The agents are given tools for data discovery and data analysis on DuckDB or Python in an isolated sandbox on an AWS \texttt{g6.2xlarge} instance (8 vCPUs, 32 GiB RAM, NVIDIA L4 GPU), where the agent policy is orchestrated with \texttt{Strands Agent}--an open-source model-driven agent SDK that runs the reason--act--observe tool-use loop given a model, system prompt, and tools. Each tool has a 150s timeout.


\stitle{Baseline difference from LakeQA} All modes use a summarizing conversation manager for context compaction as EQA tasks are often long-winded and have large intermediate query results.  We also install a plugin that nudges the agent to reconsider its current strategy when the agent has repeated similar operations over \(7\) turns.
Lastly, search tool returns the sources augmented with metadata defined in \Cref{subsubsec:search-ablation}. These additions make SANA baseline stronger than the original LakeQA benchmark; thus, our results should not be interpreted as a direct reproduction of LakeQA~\cite{wang2026lakeqaexploratoryqabenchmark}.



\subsection{Metrics}
\label{subsec:metrics}

Our primary task-success metric is semantic match (SM). We use an
LLM-as-a-judge to compare the submitted answer against the gold answer,
allowing equivalent answers that differ in formatting, aliases, units, or
phrasing.

To measure discovery behavior, let \(R\) be the unique sources retrieved from the search tools \(f_i\), and \(A\) the unique sources actually accessed by the agent through data-analysis tools \(g_i\). We report:
\[
\mathrm{D}_{\mathrm{ret}}=|\mathcal{D}_{\mathrm{gold}}\cap R|/|\mathcal{D}_{\mathrm{gold}}|,
\qquad
\mathrm{D}_{\mathrm{acc}}=|\mathcal{D}_{\mathrm{gold}}\cap A|/|\mathcal{D}_{\mathrm{gold}}|, \]
the \emph{retrieval recall} and \emph{access recall} over the gold sources. We additionally track tool-call counts and failures in each run's log.

\section{Results \& Discussion}
\label{sec:results}

This section first reports SANA's ablation results on LakeQA, then uses LakeQA for ablation and failure analyses. We then evaluate whether the same analysis holds on the converted KramaBench.

\subsection{LakeQA Ablation}
\label{subsec:results-lakeqa}

\subsubsection{Ablation Delta}
\label{subsubsec:lakeqa-ablation-delta}
\begin{table}[h]
  \centering
  \caption{LakeQA ablation matrix ($135$ tasks/cell).}
  \label{tab:sana-lakeqa}
  \footnotesize
  \setlength{\tabcolsep}{4pt}
  \renewcommand{\arraystretch}{1.05}
  \resizebox{\columnwidth}{!}{%
  \begin{tabular}{llllrrrrr}
    Model & Plan & Search & \shortstack{Data\\An.}
    & SM (\%) & $D_{acc}$ (\%) & $D_{ret}$ (\%)
    & \shortstack{Ret Tool\\Call}
    & \shortstack{Acc Tool\\Call} \\
    \midrule
    \multirow{7}{*}{\shortstack[l]{\texttt{gpt-5.4-}\\\texttt{nano}}}
       & \textsc{Naive} & \textsc{Ideal} & \textsc{Ideal} & 34.1 & 46.5 & 56.5 & 3.2 & 11.6 \\
       & \textsc{Standard} & \textsc{Ideal} & \textsc{Ideal} & 31.1 & 47.0 & 57.6 & 4.1 & 13.2 \\
       & \textsc{Ideal} & \textsc{Naive} & \textsc{Ideal} & 23.0 & 39.0 & 50.8 & 5.8 & 12.3 \\
       & \textsc{Ideal} & \textsc{Standard} & \textsc{Ideal} & 26.7 & 43.4 & 55.4 & 4.7 & 13.2 \\
       & \textsc{Ideal} & \textsc{Ideal} & \textsc{Standard} & 28.9 & 44.0 & 56.9 & 3.8 & 13.6 \\
       & \textsc{Ideal} & \textsc{Ideal} & \textsc{Ideal} & \textbf{37.0} & \textbf{47.1} & 58.1 & 4.1 & 12.5 \\
       & \textsc{Ideal} & \textsc{Preloaded} & \textsc{Ideal} & \textbf{51.8} & \textbf{59.2} & -- & 0.0 & 17.1 \\
    \midrule
    \multirow{7}{*}{\shortstack[l]{\texttt{gpt-5-}\\\texttt{mini}}}
       & \textsc{Naive} & \textsc{Ideal} & \textsc{Ideal} & 66.7 & 55.1 & 59.0 & 5.8 & 13.1 \\
       & \textsc{Standard} & \textsc{Ideal} & \textsc{Ideal} & 66.7 & 56.2 & 58.8 & 5.8 & 14.4 \\
       & \textsc{Ideal} & \textsc{Naive} & \textsc{Ideal} & 63.0 & 50.7 & 56.5 & 5.9 & 15.4 \\
       & \textsc{Ideal} & \textsc{Standard} & \textsc{Ideal} & 61.5 & 52.4 & 58.0 & 5.8 & 15.9 \\
       & \textsc{Ideal} & \textsc{Ideal} & \textsc{Standard} & 57.8 & 53.8 & 58.7 & 5.7 & 12.9 \\
       & \textsc{Ideal} & \textsc{Ideal} & \textsc{Ideal} & \textbf{76.3} & \textbf{56.5} & 60.2 & 5.9 & 14.1 \\
       & \textsc{Ideal} & \textsc{Preloaded} & \textsc{Ideal} & \textbf{77.0} & \textbf{61.7} & -- & 0.0 & 16.8 \\
  \end{tabular}
  }
\end{table}

\begin{figure*}[h]
  \centering
  \includegraphics[width=0.96\textwidth,keepaspectratio]{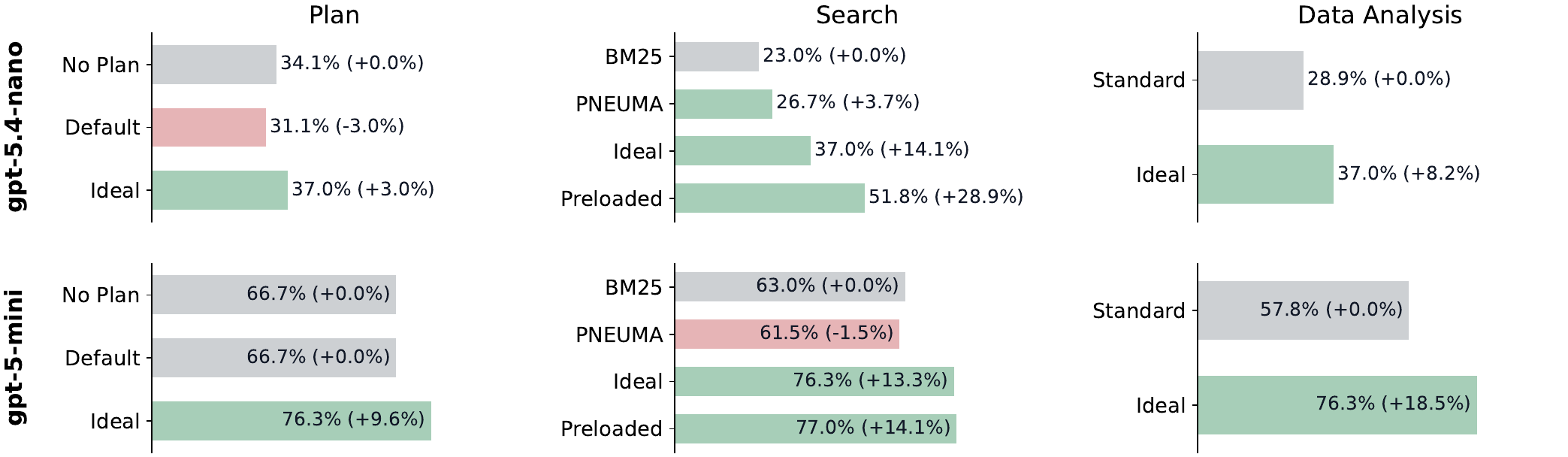}
  \vspace{-0.5em}
  \caption{LakeQA ablation semantic-match delta.}
  \Description{A two-row by three-column grid of horizontal bar charts; rows are models and columns are Plan, Search, and Data Analysis ablations for LakeQA. 
  }
  \label{fig:lakeqa-axis-delta}
  \vspace{-0.75em}
\end{figure*}
Figure~\ref{fig:lakeqa-axis-delta} reports LakeQA ablation semantic-match deltas.
Data discovery is a major bottleneck: Ideal search improves over BM25 by +14.1\% for gpt-5.4-nano and +13.3\% for gpt-5-mini. 
The BM25--Pneuma comparison is mixed: Pneuma helps gpt-5.4-nano (+3.7\%) but hurts gpt-5-mini (-1.5\%), suggesting that non-ideal search improvements remain unreliable. Preloaded Sources is the strongest SANA intervention, where relative to BM25, this yields a +28.9\% gain for gpt-5.4-nano and +14.1\% for gpt-5-mini. This suggests that weaker models are particularly bottlenecked by search navigation.

Ideal data analysis also yields large gains, especially for gpt-5-mini (+18.5\% versus +8.2\% for gpt-5.4-nano). This suggests that gpt-5-mini more often reaches the right computation intent but fails during implementation or execution, while gpt-5.4-nano more often fails by requesting the wrong operations.

Planning gains are smaller. Ideal planning improves gpt-5.4-nano by +3.0\% and gpt-5-mini by +9.6\%, suggesting that reasoning chains scale with the model's capability, discussed in Section~\ref{subsubsec:planning-axis}. Standard planning gives almost no gain, indicating that self-written plans can introduce unnecessary steps or structure that do not necessarily prevent drift or wrong-source navigation.

\subsubsection{End-to-End Mode Comparison}

\begin{table}[h]
  \centering
  \scriptsize
  \setlength{\tabcolsep}{4pt}
  \renewcommand{\arraystretch}{1.05}
  \caption{LakeQA end-to-end mode comparison ($n=135$).}
  \label{tab:lakeqa-canonical-modes}
  \begin{tabular}{@{}rlrrrrr@{}}
    \textbf{Model} & \textbf{Mode} & \textbf{SM (\%)} & \textbf{$D_{ret}$ (\%)} & \textbf{$D_{acc}$ (\%)}  & \shortstack{Ret Tool\\Call}
    & \shortstack{Acc Tool\\Call}  \\
    \midrule
    \multirow{3}{*}{\shortstack[l]{\texttt{gpt-5.4-}\\\texttt{nano}}} & \textsc{Naive} & 20.7 & 45.4 & 30.7 & 5.0 & 11.0 \\
     & \textsc{Standard} & 19.3\,\dbad{($-1.5$)} & 53.2\,\dgood{($+7.8$)} & 39.5\,\dgood{($+8.8$)} & 5.2 & 14.1 \\
     & \textsc{Ideal} & \textbf{37.0}\,\dgood{($+16.3$)} & \textbf{58.1}\,\dgood{($+12.7$)} & \textbf{47.1}\,\dgood{($+16.4$)} & 4.1 & 12.5 \\
    \midrule
    \multirow{3}{*}{\shortstack[l]{\texttt{gpt-5-}\\\texttt{mini}}} & \textsc{Naive} & 56.3 & 53.8 & 45.5 & 7.2 & 12.1 \\
     & \textsc{Standard} & 57.8\,\dgood{($+1.5$)} & 56.9\,\dgood{($+3.1$)} & 48.5\,\dgood{($+3.0$)} & 6.0 & 13.2 \\
     & \textsc{Ideal} & \textbf{76.3}\,\dgood{($+20.0$)} & \textbf{60.2}\,\dgood{($+6.4$)} & \textbf{56.5}\,\dgood{($+11.0$)} & 5.9 & 14.1 \\
  \end{tabular}

\end{table}

Table~\ref{tab:lakeqa-canonical-modes} compares the three end-to-end modes: Naive, Standard, and Ideal. Standard provides almost no gain over Naive, giving gpt-5-mini +1.5\% and giving gpt-5.4-nano -1.5\% even when \(D_{ret}\) and \(D_{acc}\) are both improved. In contrast, Ideal substantially improves both models, raising gpt-5.4-nano from 20.7\% to 37.0\% SM and gpt-5-mini from 56.3\% to 76.3\% SM. This gap shows that stronger non-ideal components do not fully remove the main EQA bottlenecks even when more data is accessed and retrieved. 
\subsection{LakeQA Ablation Analysis}

\subsubsection{Planning Ablation}
\label{subsubsec:planning-axis}
Planning Idealization produces smaller gains than search or data-analysis Idealization on LakeQA. To understand why, we separate two questions: whether the agent can write a reasonable plan, and whether it actually follows that
plan during the trajectory.

\stitle{Plan Similarity}
We first compare the plans produced under Standard and
Ideal planning. A \texttt{gpt-5.4-mini} judge labels a Standard plan as similar to the Ideal plan when it preserves the same main subquestions, source path, dependency order, and core computation. By this measure, Standard plans are often close to the Ideal scaffold: 81.5\% of \texttt{gpt-5.4-nano} plans and 77.8\% of gpt-5-mini plans are marked similar to their corresponding Ideal plans. This suggests that agents can already create plans covering the right subquestions and intents. 

\stitle{Plan Trajectory} To check whether the agent actually follows their plan, we audit each trajectory against the sanitized subquestion sequence. A \texttt{gpt-5.4-mini} judge analyzes each run's agent tool-call trajectory, labelling a trajectory as \textsc{Followed} when the agent completes the intended ordered subproblems and \textsc{Mostly Followed} when the trajectory preserves the main structure but skips or detours around a minor step. 

\begin{table}[h]
\centering
\footnotesize
\setlength{\tabcolsep}{2.5pt}
\renewcommand{\arraystretch}{0.95}

\caption{LakeQA plan trajectory audit over targeted plan ablations.}
\label{tab:planning-trajectory-audit}
\begin{tabular}{rlrrr}
 \textbf{Agent} & \textbf{Plan} & \textbf{Followed} & \textbf{Followed+Mostly} \\
 \midrule
   \multirow{3}{*}{gpt-5-mini} & Naive & 21.2\% & 49.6\% \\
   & Standard & 12.6\% & 46.4\%\,\dbad{($-3.2$)} \\
   & Ideal & 25.9\% & 56.5\%\,\dgood{($+6.9$)} \\
   \midrule
   \multirow{3}{*}{\texttt{gpt-5.4-nano}} & Naive & 5.7\% & 19.0\% \\
   & Standard & 8.4\% & 24.0\%\,\dgood{($+4.9$)} \\
   & Ideal & 7.9\% & 28.1\%\,\dgood{($+9.1$)} \\

\end{tabular}

\end{table}

The trajectory audit clarifies why ideal planning has limited impact. 
gpt-5.4-nano is especially poor at following plans: only 19.0\% of trajectories are at least \textsc{Mostly Followed} at baseline, and ideal only raises it to 28.1\%. gpt-5-mini follows plans more often, but the gain from ideal planning is still modest (49.6\% to 56.5\%).  Thus, simply providing the correct decomposition does not mean agents follow that decomposition reliably. These results suggest that agents are more bottlenecked by their ability to follow plans than their ability to decompose questions; stronger runtime scaffolds for tracking progress and recovering when their trajectory drifts from the plan are needed.

\subsubsection{Search Ablation}
\label{subsubsec:search-axis}

Figure~\ref{fig:lakeqa-axis-delta} separates two search-related bottlenecks: retrieval quality and search navigation. Ideal search improves over BM25 for
both models, showing that retrieving the right sources from a large data lake is a major bottleneck. However, Preloaded Sources improves over Ideal Search substantially on gpt-5.4-nano while barely affecting gpt-5-mini. This suggests that weaker agents also struggle at search navigation: formulating the right search queries, deciding when to search, and continuing using the right sources once they are available. In contrast, the stronger model gains little improvement from removing search entirely.

Search ablation therefore exposes two separate needs: better retrieval infrastructure and better agent policies for query formulation, source commitment, and evidence collection.

\subsubsection{Data Analysis Ablation}
\label{subsubsec:execution-axis}

Ideal data analysis helps gpt-5-mini more than gpt-5.4-nano on LakeQA (+18.5\% vs.\ +8.2\% in Figure~\ref{fig:lakeqa-axis-delta}). This suggests that gpt-5-mini often reaches the right source context and computation intent, but fails to implement the operation correctly in SQL or Python. The smaller gain for gpt-5.4-nano does not mean that execution is less important for weaker models: rather, many gpt-5.4-nano failures occur before data analysis implementation becomes a bottleneck: choosing the wrong source, scope, or analysis intent. Since ideal execution preserves the agent's stated intent, it fixes implementation errors but not wrong requests. 
Thus, data analysis ablation mainly separates code-generation failures from upstream source-selection and intent errors.

\subsection{LakeQA Failure Analysis}
\label{subsec:failure-analysis}

In this section, we audit failed traces and categorize where every evaluation run with an incorrect semantic outcome fails.  

\stitle{Audit Method} Each failed run is reviewed using a two-stage audit with  \texttt{gpt-5.4-mini}. First, the auditor reads the task, the SANA profile, evaluation results, and log trace, then extracts a compact evidence trace of all failures and the turns at which they occur. Then, a separate labeler assigns one or more failure events from the taxonomy in \Cref{tab:answer-failure-groups} using only that evidence trace, such that a single failed run may contribute multiple failures such as answer finalization and planning failures. Since co-occurring symptoms are recorded together regardless of root cause (e.g., misreading question constraint and then filtering the wrong rows returns both constraint misread and wrong scope/filter labels), we treat these LLM-assisted labels as diagnostic rather than ground truth.
\begin{table}[t]
\centering
\scriptsize
\setlength{\tabcolsep}{1.7pt}
\renewcommand{\arraystretch}{1.03}
\caption{Failure families in LakeQA traces. Entries are event shares; Meaning lists (gpt-5-mini, gpt-5.4-nano) subgroups.}
\label{tab:answer-failure-groups}
\begin{tabularx}{\columnwidth}{p{0.23\columnwidth}p{0.12\columnwidth}p{0.14\columnwidth}X}
Group & gpt-5-mini & gpt-5.4-nano & Meaning \\
\midrule
Task/planning failures & 7.4\% & 11.3\% &  Reasoning chain divergence (5.4\%, 8.7\%); question constraint misread (1.9\%, 2.6\%). \\
Wrong source target failures & 0.0\% & 7.6\% & Chose the wrong dataset, table, source family, or source version. \\
Execution/computation failures & 39.6\% & 26.2\% & Computation intent failures: wrong scope/filter (30.9\%, 20.1\%); computation/aggregation error (4.9\%, 2.8\%); extraction/parsing error (3.9\%, 3.2\%). \\
Incomplete evidence failures & 12.2\% & 12.3\% & Ran out of budget before gathering enough evidence (10.3\%, 5.0\%); submitted early with incomplete evidence (1.9\%, 7.3\%). \\
Turn-waste failures & 2.1\% & 8.8\% &
Repeated actions without progress: query execution/repair loop (0.8\%, 5.5\%); schema inspection loop (0.2\%, 1.8\%); low-yield search loop (0.6\%, 1.4\%); same-hop repetition even after gathering enough evidence (0.6\%, 0.2\%). \\
Finalization failures & 21.0\% & 13.2\% & Correct evidence or computed value appeared, but the submitted answer was wrong. \\
Tool blocker failures & 17.7\% & 20.6\% & Files, tools, repair calls, unsupported formats, or runtime limits blocked progress. \\
\end{tabularx}
\vspace{-0.75em}
\end{table}

\begin{figure}[h]
  \centering
  \includegraphics[width=\linewidth, height = 13em]{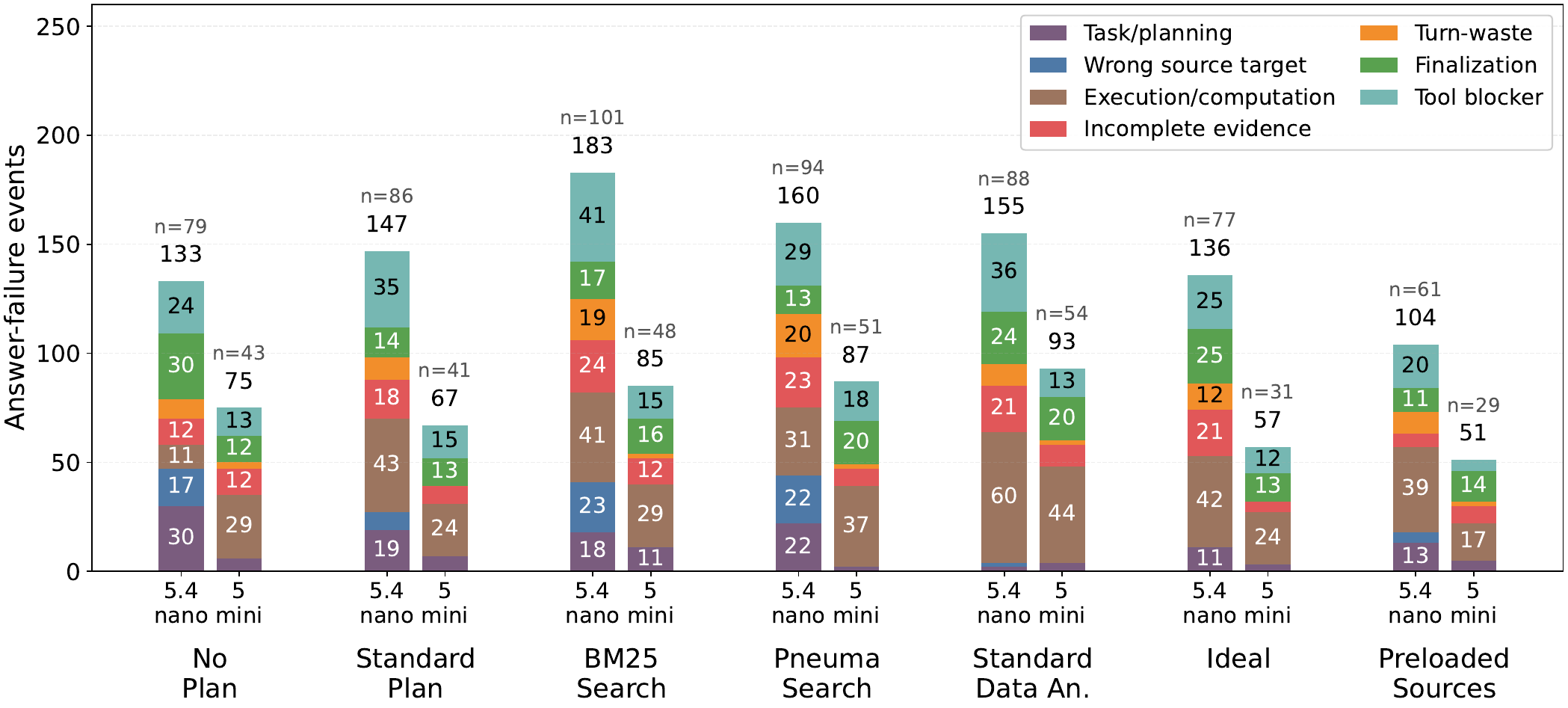}
  \vspace{-0.25em}
  \caption{Answer-failure groups across LakeQA ablations. Bold numbers are total events and \(n\) the failed runs per bar.}
  \Description{A stacked vertical bar chart with SANA conditions on the horizontal axis, two model-specific bars within each condition, and colored answer-failure group segments stacked within each model bar.}
  \label{fig:answer-failure-families}
  \vspace{-0.75em}
\end{figure}

Figure~\ref{fig:answer-failure-families} and Table~\ref{tab:answer-failure-groups} explain why the two models respond differently to the same conditions. Gpt-5.4-nano produced 1018 failure events over 586 failed runs and gpt-5-mini 515 over 297, about 1.74 events per failed run for both models. Table~\ref{tab:answer-failure-groups} gives the event distribution within each model and Figure~\ref{fig:answer-failure-families} the condition-level counts. The distributions differ: gpt-5.4-nano's failures spread across all groups, while gpt-5-mini's concentrate in execution/computation (57.2\% of its failed runs) and finalization (35.7\%).

The clearest gap is source scoping: gpt-5-mini has 0\% wrong-source failure labels whereas gpt-5.4-nano receives 12.8\%. This matches \cref{fig:lakeqa-axis-delta}, where gpt-5.4-nano benefits more from Ideal Search and Preloaded Sources as they prevent wrong-source branches. Gpt-5.4-nano's other disproportionate failures---turn-waste (8.8\% gpt-5.4-nano vs.\ 2.1\% gpt-5-mini), reasoning-chain divergence (8.7\% vs.\ 5.4\%), and early submission with incomplete evidence (7.3\% vs.\ 1.9\%)---are trajectory failures, agreeing with Subsection~\ref{subsubsec:planning-axis} where gpt-5.4-nano is worse at staying on its plan.

Idealizing search (BM25 to Ideal in Figure~\ref{fig:answer-failure-families}) reduces gpt-5.4-nano's wrong-source (23 to 0), turn-waste (19 to 12), and divergence (18 to 11) events, partly because failed runs themselves drop (101 to 77). For gpt-5-mini these are already near zero and idealized search mainly trims incomplete-evidence errors, suggesting the stronger model already selects sources and operations well while the weaker one still struggles.

Planning is more selective, helping mainly the smaller model read question constraints and finalize answers. Finalization, computation intent, and turn budget remain failure modes for both, even under Ideal and Preloaded where source access is solved. Overall the two models fail differently: gpt-5-mini usually reaches the right evidence but mis-executes or mis-finalizes, while gpt-5.4-nano's failures spread across source targeting, plan divergence, turn-waste, and premature stopping, reflecting weaker trajectory control. The ablation gains thus reflect differences in action policies, not only component quality.

\begin{figure*}[t]
  \centering
  \includegraphics[width=0.96\textwidth,keepaspectratio]{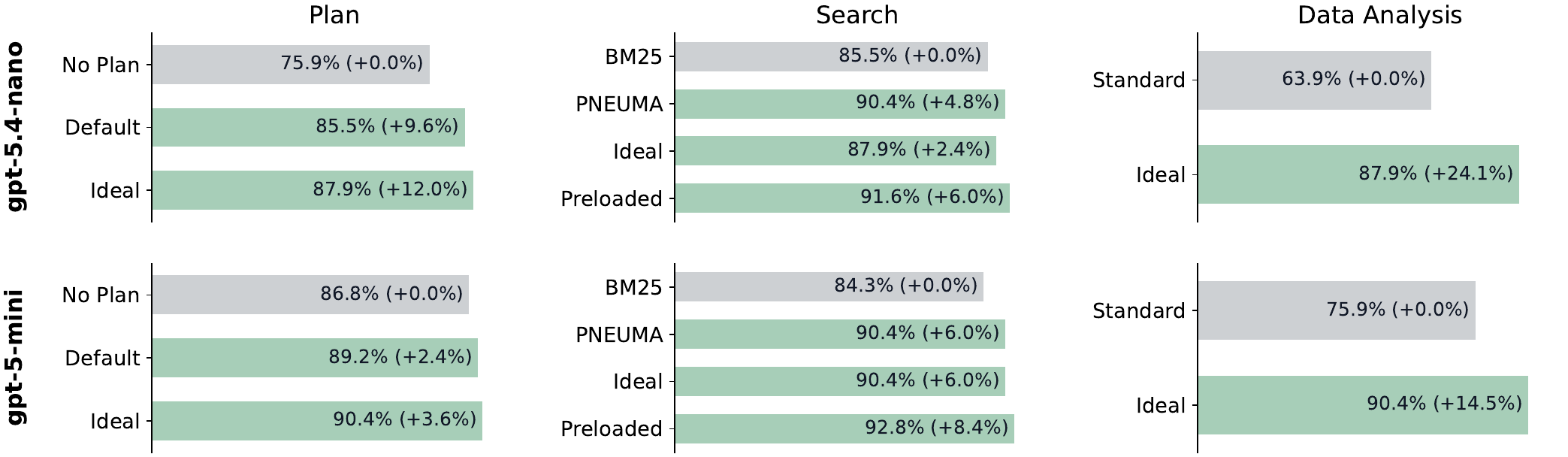}
  \vspace{-0.5em}
  \caption{KramaBench ablation semantic-match delta.}
  \Description{A two-row by three-column grid of horizontal bar charts; rows are models and columns are Plan, Search, and Data Analysis ablations for KramaBench.}
  \label{fig:krama-axis-delta}
  \vspace{-0.75em}
\end{figure*}

\subsection{KramaBench Ablation}
\label{subsec:results-krama}

\subsubsection{Ablation Delta}

\begin{table}[h]
  \centering
  \caption{KramaBench ablation matrix ($83$ tasks/cell).}
  \label{tab:sana-krama}
  \footnotesize
  \setlength{\tabcolsep}{4pt}
  \renewcommand{\arraystretch}{1.05}
  \resizebox{\columnwidth}{!}{%
  \begin{tabular}{llllrrrrr}
    Model & Plan & Search & \shortstack{Data\\An.}
    & SM (\%) & $D_{acc}$ (\%) & $D_{ret}$ (\%)
    & \shortstack{Ret Tool\\Call}
    & \shortstack{Acc Tool\\Call} \\
    \midrule
    \multirow{7}{*}{\shortstack[l]{\texttt{gpt-5.4-}\\\texttt{nano}}}
       & \textsc{\textsc{Naive}} & \textsc{\textsc{Ideal}} & \textsc{\textsc{Ideal}} & 75.9 & 81.7 & 89.9 & 1.7 & 3.2 \\
       & \textsc{Standard} & \textsc{\textsc{Ideal}} & \textsc{\textsc{Ideal}} & 85.5 & 89.3 & 91.0 & 1.4 & 3.7 \\
       & \textsc{\textsc{Ideal}} & \textsc{\textsc{Naive}} & \textsc{\textsc{Ideal}} & 85.5 & 50.1 & 78.9 & 2.9 & 4.0 \\
       & \textsc{\textsc{Ideal}} & \textsc{Standard} & \textsc{\textsc{Ideal}} & 90.4 & 43.5 & 77.1 & 2.1 & 4.0 \\
       & \textsc{\textsc{Ideal}} & \textsc{\textsc{Ideal}} & \textsc{Standard} & 63.9 & 92.9 & 93.9 & 1.3 & 2.9 \\
       & \textsc{\textsc{Ideal}} & \textsc{\textsc{Ideal}} & \textsc{\textsc{Ideal}} & \textbf{87.9} & \textbf{92.3} & 95.8 & 1.6 & 3.5 \\
       & \textsc{\textsc{Ideal}} & \textsc{Preloaded} & \textsc{\textsc{Ideal}} & \textbf{91.6} & \textbf{98.4} & -- & 0.0 & 3.7 \\
    \midrule
    \multirow{7}{*}{\shortstack[l]{\texttt{gpt-5-}\\\texttt{mini}}}
       & \textsc{\textsc{Naive}} & \textsc{\textsc{Ideal}} & \textsc{\textsc{Ideal}} & 86.8 & 93.9 & 95.8 & 1.4 & 4.6 \\
       & \textsc{Standard} & \textsc{\textsc{Ideal}} & \textsc{\textsc{Ideal}} & 89.2 & 93.7 & 94.3 & 1.4 & 4.6 \\
       & \textsc{\textsc{Ideal}} & \textsc{\textsc{Naive}} & \textsc{\textsc{Ideal}} & 84.3 & 51.6 & 82.0 & 2.7 & 4.2 \\
       & \textsc{\textsc{Ideal}} & \textsc{Standard} & \textsc{\textsc{Ideal}} & 90.4 & 50.8 & 78.9 & 2.0 & 4.1 \\
       & \textsc{\textsc{Ideal}} & \textsc{\textsc{Ideal}} & \textsc{Standard} & 75.9 & 96.7 & 96.7 & 1.5 & 3.4 \\
       & \textsc{\textsc{Ideal}} & \textsc{\textsc{Ideal}} & \textsc{\textsc{Ideal}} & \textbf{90.4} & \textbf{97.0} & 97.6 & 1.2 & 4.1 \\
       & \textsc{\textsc{Ideal}} & \textsc{Preloaded} & \textsc{\textsc{Ideal}} & \textbf{92.8} & \textbf{99.0} & -- & 0.0 & 4.6 \\
  \end{tabular}
  }
\end{table}

Figure~\ref{fig:krama-axis-delta} shows a different profile from LakeQA. Data analysis is the largest KramaBench intervention, improving gpt-5.4-nano by +24.1\% and gpt-5-mini by +14.5\%, indicating that KramaBench is analysis-heavy. Search is less limiting: Ideal search improves over BM25 by +2.4\% for gpt-5.4-nano and +6.0\% for gpt-5-mini, while Preloaded Sources adds +6.0\% and +8.4\%.  This likely reflects the benchmark construction: retrieval is artificially introduced by converting KramaBench into a LakeQA-style task, while the underlying tasks remain centered on data-to-insight execution. Planning helps gpt-5.4-nano more than gpt-5-mini, suggesting that weaker models benefit more from explicit decomposition on KramaBench. 

\subsubsection{End-to-End Mode Comparison}

\begin{table}[h]
  \centering
  \scriptsize
  \setlength{\tabcolsep}{4pt}
  \renewcommand{\arraystretch}{01.05}
    \caption{KramaBench end-to-end mode comparison ($n=83$).}
    \label{tab:krama-canonical-modes}
  \begin{tabular}{@{}llrrrrr@{}}
    \textbf{Model} & \textbf{Mode} & \textbf{SM (\%)} & \textbf{$D_{ret}$ (\%)} & \textbf{$D_{acc}$ (\%)} & \shortstack{Ret Tool\\Call}
    & \shortstack{Acc Tool\\Call}  \\
    \midrule
    \multirow{3}{*}{\shortstack[l]{\texttt{gpt-5.4-}\\\texttt{nano}}} & \textsc{Naive} & 44.6 & 64.2 & 31.8 & 3.2 & 3.3 \\
     & \textsc{Standard} & 57.8\,\dgood{($+13.3$)} & 70.6\,\dgood{($+6.4$)} & 44.6\,\dgood{($+12.8$)} & 2.4 & 3.6 \\
     & \textsc{Ideal} & \textbf{87.9}\,\dgood{($+43.4$)} & \textbf{95.8}\,\dgood{($+31.6$)} & \textbf{92.3}\,\dgood{($+60.4$)} & 1.6 & 3.5 \\
     \midrule
    \multirow{3}{*}{\shortstack[l]{\texttt{gpt-5-}\\\texttt{mini}}} & \textsc{Naive} & 62.6 & 64.7 & 36.0 & 3.8 & 3.6 \\
     & \textsc{Standard} & 66.3\,\dgood{($+3.6$)} & 68.5\,\dgood{($+3.9$)} & 34.8\,\dbad{($-1.2$)} & 3.2 & 3.8 \\
     & \textsc{Ideal} & \textbf{90.4}\,\dgood{($+27.7$)} & \textbf{97.6}\,\dgood{($+32.9$)} & \textbf{97.0}\,\dgood{($+61.1$)} & 1.2 & 4.1 \\
  \end{tabular}

\end{table}

\Cref{tab:krama-canonical-modes} shows that Standard improves over Naive on KramaBench, especially for gpt-5.4-nano (+13.3\%). Ideal produces the largest gains, reaching 87.9\% SM for nano and 90.4\% SM for gpt-5-mini. Under Ideal, retrieval recall reaches 95.8\%--97.6\% and access recall reaches 92.3\%--97.0\%, so the remaining errors are unlikely to come mainly from missing sources.

\section{Conclusion and Limitations}
\label{sec:conclusion}

\stitle{Conclusion}
SANA asks a diagnostic question: when an EQA agent fails over a data lake, which part of the runtime is responsible? We answer this by varying only the planning, search, and data-analysis interfaces. Across LakeQA and the converted KramaBench subset, the results show that EQA failures are not explained by a single component. Data analysis is a consistent bottleneck, search becomes a major limitation in large-scale discovery settings, and residual policy failures remain even after component idealization.

On LakeQA, search is a major bottleneck. Ideal search improves over BM25 for both agents, and preloading gold sources further improves the weaker agent, showing that LakeQA stresses both retrieval quality and search navigation: formulating useful queries, deciding when to search, and committing to the right source context. Data analysis is also critical because agents that reach relevant sources can still fail to implement its intent. Planning is less limiting: agents often produce plans close to the gold decomposition, but they do not reliably follow those plans.


The converted KramaBench has a different bottleneck. Because its tasks contain fewer gold sources and remain closer to data-to-insight analysis, search is less dominant, while data-analysis becomes the main bottleneck. This shows that SANA does not rank search, planning, or execution as universally hardest, but exposes how bottlenecks change with the benchmark structure.

On LakeQA, the failure audit helps explain the residual gap left after component idealization. Smaller agents fail broadly across source choice, progress control, incomplete evidence, tool blockers, and finalization. Stronger agents are better at staying within the right source context, but still fail during computation and final-answer synthesis. These results suggest that future EQA systems need more than better retrievers or stronger code generation. They need runtime policies that track subgoal progress, commit to validated sources, recover from trajectory drifts, verify intermediate evidence, and check the final answer before stopping.


\stitle{Limitations}
\textbf{(i). Diagnostic idealization.} SANA's ideal tools are diagnostic approximations of the oracles. Ideal search selects only from \(\mathcal{D}_{\mathrm{gold}}\), so alternative valid sources outside the profile are not credited. Similarly, ideal data analysis relies on semantically matching against annotated records; false matches or missed matches may happen. Thus, SANA's ablations should be interpreted as approximate diagnostic upper bounds. 
\textbf{(ii). Task-suite scope.} The results are tied to the evaluated task suites. We evaluate LakeQA and a LakeQA-converted KramaBench. The conclusions should therefore be read as evidence about these data-lake settings, not as a universal ranking of search, planning, and data analysis difficulty across all agent benchmarks. 
\textbf{(iii). Judge-based measurements.} Several measurements rely on LLM judges, including semantic-match, plan-trajectory assessment, and failure-audit labeling. These judges make the evaluation practical and allow semantically equivalent answers to be credited, but they can introduce false positives, false negatives, or taxonomy-dependent failure assignments.


\clearpage

\bibliographystyle{ACM-Reference-Format}
\bibliography{references}

\end{document}